\documentclass[conference]{IEEEtran}
\IEEEoverridecommandlockouts
\usepackage{cite}
\usepackage{amsmath,amssymb,amsfonts}
\usepackage{algorithmic}
\usepackage{graphicx}
\usepackage{textcomp}
\usepackage{xcolor}
\usepackage{float}
\usepackage{multirow}
\usepackage{booktabs}
\def\BibTeX{{\rm B\kern-.05em{\sc i\kern-.025em b}\kern-.08em
    T\kern-.1667em\lower.7ex\hbox{E}\kern-.125emX}}
\begin{document}

\title{EG-Gaussian: Epipolar Geometry and Graph Network Enhanced 3D Gaussian Splatting}

\author{
\IEEEauthorblockN{ Beizhen Zhao, Yifan Zhou, Zijian Wang, Hao Wang*\thanks{
* Corresponding author.} }
\IEEEauthorblockA{\textit{The Hong Kong University of Science and Technology (Guangzhou)}, Guangzhou, China \\
Email: bzhao610@connect.hkust-gz.edu.cn, \\
haowang@hkust-gz.edu.cn}
}

\maketitle

\begin{abstract}
In this paper, we explore an open research problem concerning the reconstruction of 3D scenes from images. Recent methods have adopt 3D Gaussian Splatting (3DGS) to produce 3D scenes due to its efficient training process. 
However, these methodologies may generate incomplete 3D scenes or blurred multiviews. 
This is because of (1) inaccurate 3DGS point initialization and (2) the tendency of 3DGS to flatten 3D Gaussians with the sparse-view input.
To address these issues, we propose a novel framework EG-Gaussian, which utilizes epipolar geometry and graph networks for 3D scene reconstruction. Initially, we integrate epipolar geometry into the 3DGS initialization phase to enhance initial 3DGS point construction. Then, we specifically design a graph learning module to refine 3DGS spatial features, in which we incorporate both spatial coordinates and angular relationships among neighboring points. 
Experiments on indoor and outdoor benchmark datasets demonstrate that our approach significantly improves reconstruction accuracy compared to 3DGS-based methods. 

\end{abstract}

\begin{IEEEkeywords}
3D Reconstruction, Gaussian Splatting, Epipolar Geometry, Graph Neural Network
\end{IEEEkeywords}

\begin{figure*}
\centerline{\includegraphics[width=1.0\textwidth]{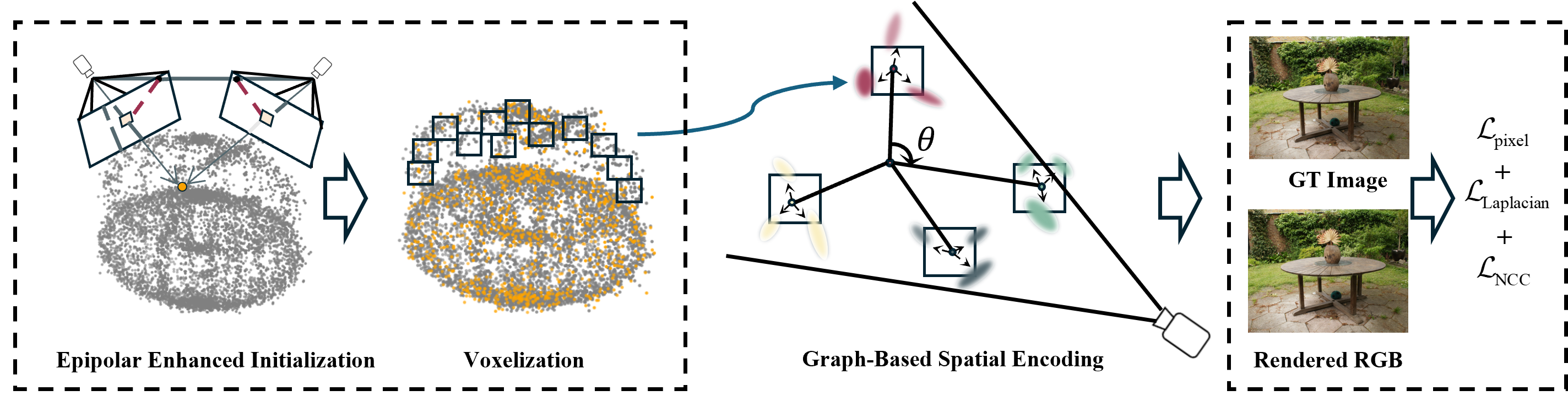}}
\caption{\textbf{Overview of EG-Gaussian.} We begin by initializing the 3D points using epipolar geometry to improve their accuracy. Next, we voxelize the 3D points, associating each voxel with multiple Gaussians. A graph-based module is then employed to model the spatial correlations between the nodes. Finally, a composite loss function is computed, incorporating pixel-based loss, NCC loss, and Laplacian pyramid loss, which are evaluated through the rendered RGB images and the original images.}
\vspace{-10pt}
\label{fig0}
\end{figure*}

\section{Introduction}
\label{sec:introduction}
3D scene reconstruction from multi-view images have garnered substantial attention due to its critical applications in fields such as virtual reality, autonomous driving and robotics. A variety of techniques have been developed to address this task, spanning from traditional methods, including structure-from-motion (SfM) \cite{schonberger2016structure} and Multi-view Stereo (MVS) \cite{seitz2006comparison}, to learning-based approaches, such as Neural Radiance Fields (NeRF) \cite{mildenhall2020nerf} and 3D Gaussian Splatting (GS) \cite{kerbl20233d}. 

Specifically, traditional methods are highly dependent on handcraft feature point matching and geometric constraints discovery within images. Additionally, they often struggle to obtain dense and precise depth estimations in textureless or highly reflective regions. In contrast, learning-based approaches adopt volumetric representations for 3D scene modeling, which obtains impressive results in terms of visual quality and scene synthesis.

Recently, 3D Gaussian Splatting \cite{kerbl20233d, yu2024mip, wu20244d} has shown superior performance. By optimizing the properties of the explicit 3D Gaussians and integrating rasterization techniques, this approach has achieved remarkable improvements in both training and rendering speeds, compared to NeRF-based methods. 
For example, Scaffold-GS \cite{chen2024pgsr} enhances structural priors and reduces memory overhead by introducing Gaussian voxelization. Building upon Scaffold-GS, Octree-GS\cite{ren2024octree} further incorporates an octree structure to improve model scalability. 

However, these approaches heavily rely on heuristic methods for 3DGS point initialization, which often fail in the presence of noisy or sparse correspondences. 
Moreover, these methods also fail to address the long-range dependencies between points in the scene. Therefore, they often struggle to capture the fine-grained spatial details, which are crucial for accurate reconstruction.

To address the limitations above, we introduce a novel EG-Gaussian framework that aims at enhancing the 3DGS initialization and feature learning phases of scene reconstruction. 
First, we leverage epipolar geometry \cite{xu2013epipolar, zhang1998determining, svoboda1998epipolar} and incorporate triangulation techniques to provide geometrically constrained 3DGS initialized points, mitigating the inherent traditional point-matching method limitations.
Secondly, to capture complex spatial dependencies among points, we integrate a graph learning module to learn spatial structure information. 
This attention mechanism enables our model to learn spatial relationships within the scene more effectively, improving reconstruction quality.
Finally, we propose an innovative spatial position encoding method that combines 3D coordinates with angular relationships, further enhancing the model’s ability to represent the geometric structure of the scene. 
Fig. \ref{fig0} illustrates the overall architecture.

Through experiments on both indoor and outdoor datasets, our method effectively restores high-frequency spatial details and accurately captures intricate textures. Our proposed EG-Gaussian produces significantly enhanced fidelity and visual quality in the reconstructed 3D scenes. 

In summary, our contributions are as follows: 

\begin{enumerate}
    \item We leverage epipolar geometry to improve the 3DGS initialization phase, ensuring the initial spatial relationships are geometrically constrained. This approach helps the model reduce the initialization ambiguities, which are often observed in traditional GS-based methods.
    
    \item We introduce a graph learning module. This method allows for more accurate feature aggregation by dynamically learning the most relevant spatial relationships among 3D points. This effectively captures both local and global scene structures.
    
    \item We propose a spatial encoding method that combines the 3D coordinates with angular information between neighboring points. This encoding significantly improves the network's ability to capture the subtle geometric nuances necessary for high-accuracy reconstruction.
\end{enumerate}

\section{Related Work}
\label{sec:related_work}

\subsection{Gaussian Splatting and Variants}
Gaussian Splatting (GS) has become a popular technique for 3D scene representation due to its efficiency and high-quality rendering performance. Methods such as \cite{kerbl20233d} have demonstrated that 3D scenes can be represented as collections of 3D Gaussians, which are more memory-efficient compared to NeRF-based representations. GS allows for continuous scene rendering by efficiently splatting Gaussians onto a 3D space. However, despite its efficiency, GS-based methods such as Scaffold-GS \cite{lu2024scaffold} and Octree-GS \cite{ren2024octree} still face challenges in spatial information integration and the accuracy of initial point placement. These methods often struggle to capture global scene dependencies, leading to suboptimal reconstructions in complex environments.

\subsection{Graph Attention Networks for Spatial Learning}
Graph neural networks (GNNs)\cite{wu2020comprehensive, zhou2020graph, scarselli2008graph} have garnered considerable attention for their ability to model intricate spatial and relational dependencies in data\cite{shi2020point}. Graph Attention Networks (GATs) \cite{velivckovic2017graph, brody2021attentive, ye2021sparse} further extend GNNs by incorporating attention mechanisms that enable the network to prioritize the most relevant neighbors for each node within a graph. This approach has proven to be particularly effective in tasks that involve capturing relationships across irregular, non-grid data structures, such as 3D point clouds. Recently, research has explored the application of GATs for 3D scene reconstruction, facilitating more accurate feature propagation and enhanced spatial understanding. However, to date, the application of GATs in Gaussian splat (GS)-based methods remains limited, with the integration of spatial dependencies between Gaussian splats posing a significant challenge.

\begin{figure}
\centerline{\includegraphics[width=0.38\textwidth]{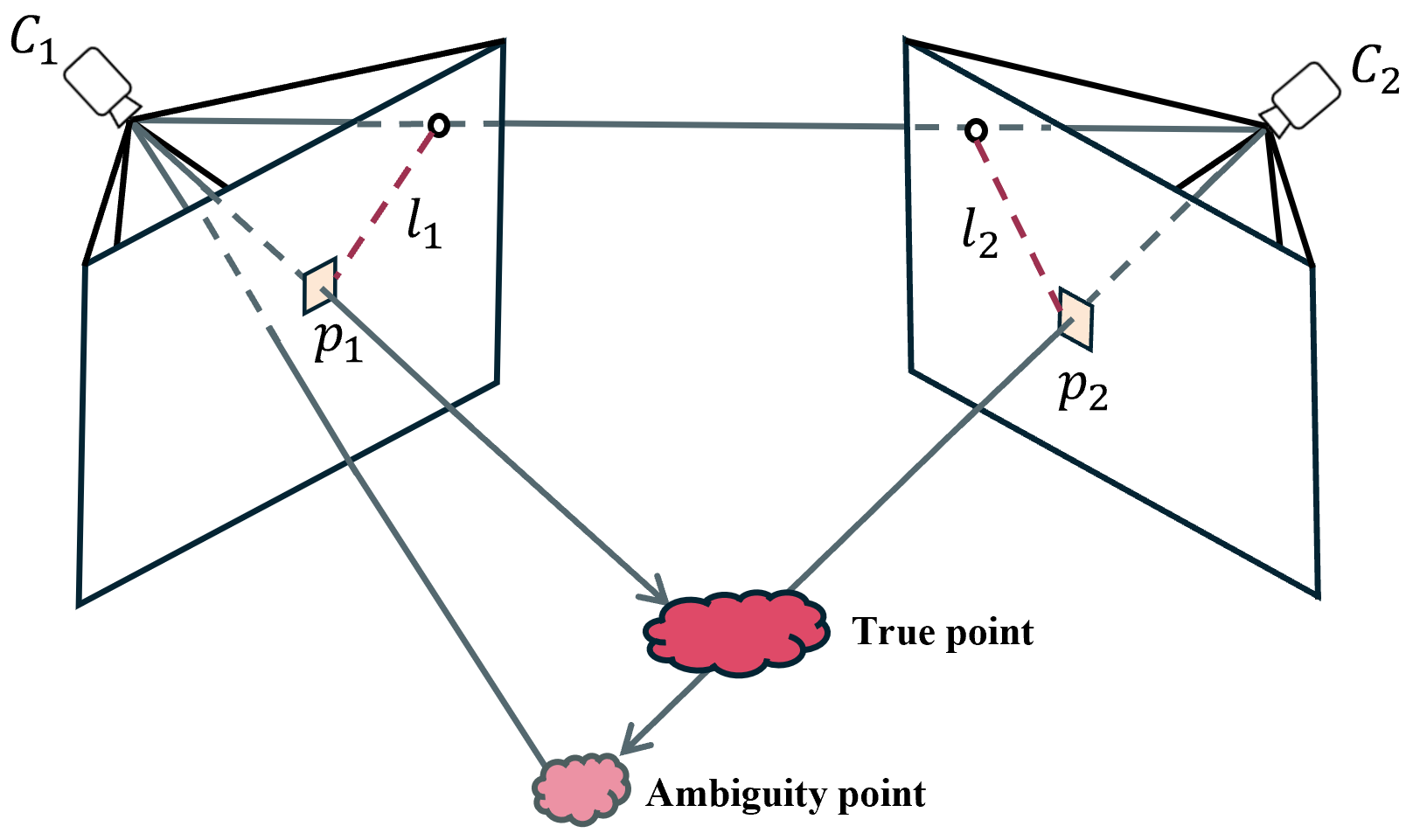}}
\caption{\textbf{Epipolar enhanced initialization.} In scenarios with sparse viewpoint constraints, the localization of 3D points may exhibit ambiguity. To mitigate this issue, we leverage epipolar geometry to establish correspondences between different camera perspectives, thereby enabling the generation of more accurate and robust 3D points.}
\vspace{-15pt}
\label{fig1}
\end{figure}

\section{Methodology}

\subsection{Epipolar Geometry for 3D Point Localization}

The first part involves using epipolar geometry for the 3D localization of points. As shown in Fig. \ref{fig1}, when given two images with corresponding 2D points \( \mathbf{x}_1 = (u_1, v_1) \) and \( \mathbf{x}_2 = (u_2, v_2) \) in the two camera views, the epipolar constraint between the two views can be represented as:

\begin{equation}
\mathbf{x}_2^\top \mathbf{F} \mathbf{x}_1 = 0,
\label{eq:epipolar_constraint}
\end{equation}
where \( \mathbf{F} \) is the fundamental matrix, and the points \( \mathbf{x}_1 \) and \( \mathbf{x}_2 \) are homogeneous coordinates of the corresponding points in the two images. \( \mathbf{F} \) can be computed as:

\begin{equation}
\mathbf{F} = (K_2^{-1})^T [t]_{\times} R K_1^{-1},
\label{eq:fundamental_matrix}
\end{equation}
where \( [t]_{\times} \) is the skew-symmetric matrix corresponding to the cross product with the translation vector, and \( K \) denotes the intrinsic matrix of the camera.

Using the fundamental matrix, the epipolar constraint ensures that the points \( \mathbf{x}_1 \) and \( \mathbf{x}_2 \) lie on corresponding epipolar lines in their respective images.
In the context of 3D reconstruction, if we have the projection matrices \( \mathbf{P}_1 \) and \( \mathbf{P}_2 \) for the two cameras, the fundamental matrix \( F \) can be derived.

Given the projection matrices \( \mathbf{P}_1 \) and \( \mathbf{P}_2 \) of the two cameras, the 3D point \( \mathbf{X} = (X, Y, Z)^\top \) can be computed by solving the system of equations:

\begin{equation}
\mathbf{P}_1 \mathbf{X} = \lambda_1 \mathbf{x}_1, \quad \mathbf{P}_2 \mathbf{X} = \lambda_2 \mathbf{x}_2,
\label{eq:triangulation_system}
\end{equation}
where \( \lambda_1 \) and \( \lambda_2 \) are scale factors. This system represents the projection of the 3D point \( \mathbf{X} \) into the 2D image coordinates \( \mathbf{x}_1 \) and \( \mathbf{x}_2 \).

To estimate the 3D point \( \mathbf{X} \) and ensure that the 3D point \( \mathbf{X} \) best satisfies the projection constraints from both views., we minimize the reprojection error by solving:

\begin{equation}
\min_{\mathbf{X}} \sum_{i=1}^{2} \left\| \mathbf{P}_i \mathbf{X} - \lambda_i \mathbf{x}_i \right\|_2^2,
\label{eq:reprojection_error}
\end{equation}

\subsection{Graph Neural Network for Feature Extraction}

\begin{figure}
\centerline{\includegraphics[width=0.45\textwidth]{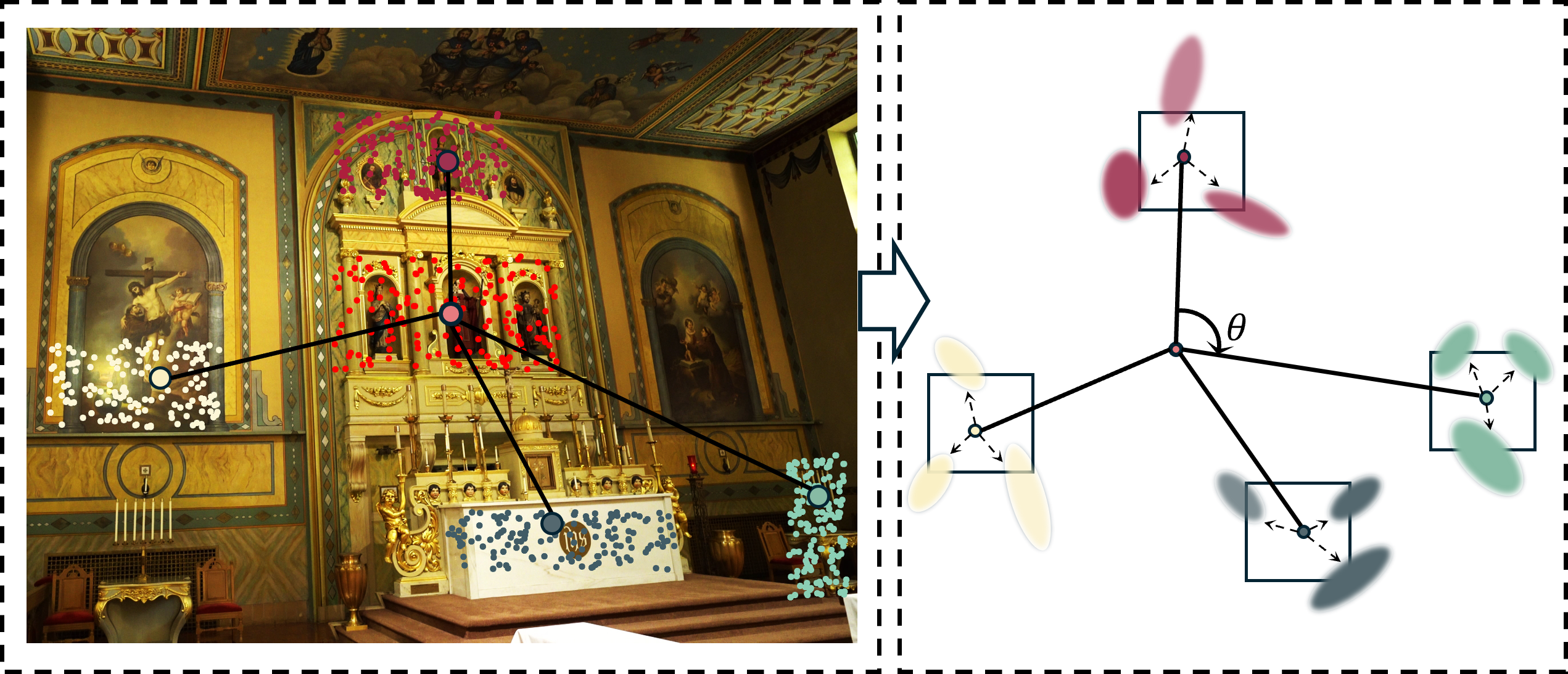}}
\caption{\textbf{Graph aggregation and spatial encode.} After voxelization, we design a graph module to capture spatial structure information. Using the 3D coordinates, we construct the neighborhood of each node, embedding the angular relationships among neighboring nodes to extract spatial structural details. A self-attention mechanism is then employed to further aggregate and refine the spatial features.}
\vspace{-15pt}
\label{fig2}
\end{figure}

Inspired by Scaffold-GS \cite{lu2024scaffold}, we initialize the 3D Gaussian by voxelizing the point cloud. Then we represent the set of initialized 3D points as a graph $\mathcal{G} = (\mathcal{V}, \mathcal{E})$, where each node $v_i \in \mathcal{V}$ corresponds to the center of a voxel, and edges $\mathcal{E}$ connect nodes that are spatially adjacent based on k-nearest neighbors, as shown in Fig. \ref{fig2}.

For each point \(  v_i \) in the cluster, we compute the pairwise distance between the point and its neighbors, and use this information to create a set of neighbor indices \( \mathbf{I} \in \mathbb{R}^{M \times k} \), which stores the indices of the \( k \) nearest neighbors for each point.

\subsubsection{Feature Aggregation}

Once the graph structure is created, we aggregate features from neighboring nodes to enhance the representation of each node. Given the feature tensor \( \mathbf{f} \), we aggregate the features of each point and its neighbors:
\begin{equation}
\mathbf{f}_{\text{agg}} = \text{concat}(\mathbf{f}, \Delta \mathbf{f}),
\label{eq:aggregated_features}
\end{equation}
where \( \mathbf{f}_{\text{agg}} \in \mathbb{R}^{M \times K \times 2F} \) denotes the aggregated feature tensor, the \( \Delta \mathbf{f} \) is calculated by:
\begin{equation}
\Delta \mathbf{f} = \mathbf{f}[\mathbf{I}] - \mathbf{f},
\label{eq:neighbor_features}
\end{equation}

\subsubsection{Spatial Encoding}

To capture the angular relationships between a node and its neighbors, we compute the angles between the vectors connecting the node to each of its neighboring points. For a node $v_i$ and its neighbor $v_j$, the angle $\theta_{ij}$ is given by:

\begin{equation}
\theta_{ij} = \arccos\left( \frac{(v_j - v_i) \cdot (v_k - v_i)}{\|v_j - v_i\| \|v_k - v_i\|} \right),
\label{eq:angular_encoding}
\end{equation}

where $v_k$ is another neighboring point of $v_i$.

These angular values are then embedded using a similar sinusoidal encoding:

\begin{equation}
\begin{split}
\gamma(\theta_{ij}) = \Bigl[ 
& \sin(2^0 \pi \theta_{ij}),\ \cos(2^0 \pi \theta_{ij}),\ \ldots, \\
& \sin(2^{L-1} \pi \theta_{ij}),\ \cos(2^{L-1} \pi \theta_{ij}) \Bigr]
\end{split}
\label{eq:angular_positional_encoding}
\end{equation}

\subsubsection{Self-Attention Mechanism}
The aggregated features are processed by a multi-head self-attention mechanism to capture both local and global context.
\begin{equation}
\mathbf{f}_{g} = \text{Multi-Head-Attention}(\mathbf{f}_{\text{agg}}, \mathbf{f}_{\text{agg}}, \mathbf{f}_{\text{agg}}, \gamma(\theta)).
\label{eq:multi_head_attention}
\end{equation}

\begin{figure}
\centerline{\includegraphics[width=0.45\textwidth]{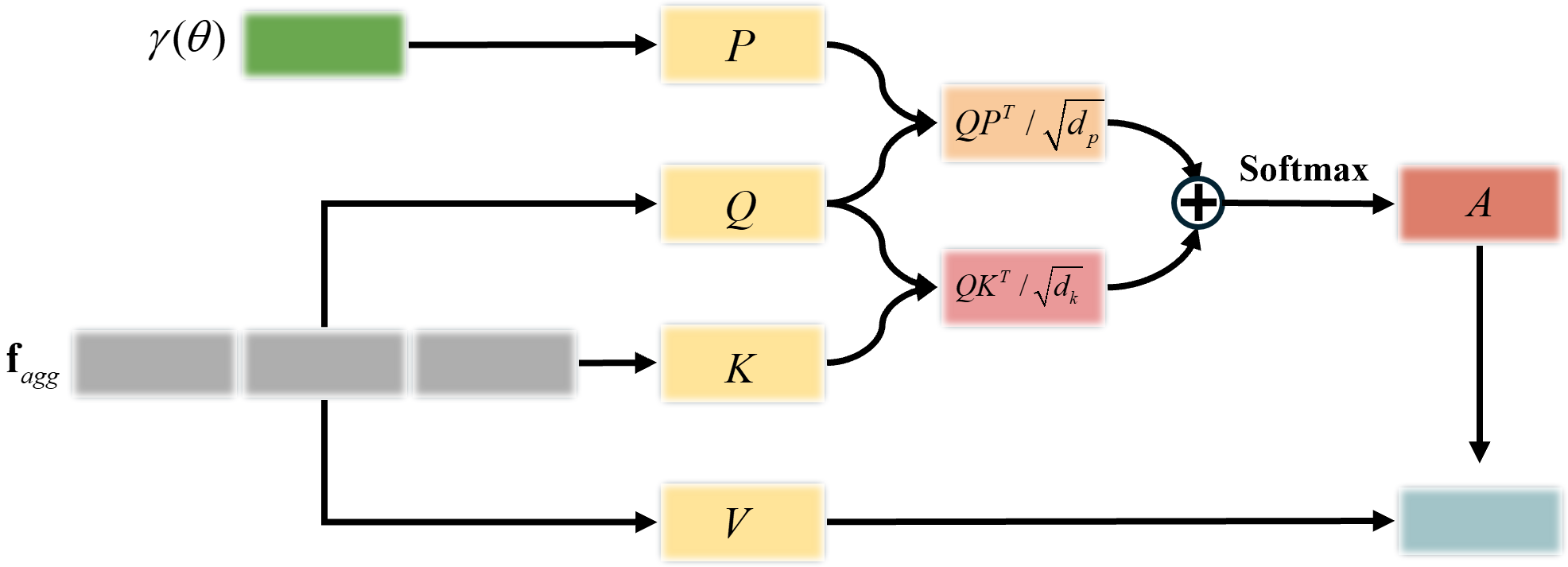}}
\caption{\textbf{Computational Flow of our Multi-Head Attention Mechanism.} The standard self-attention mechanism is extended by incorporating angular embedding information, which is fused into the attention computation to derive the final scores. More details can be found in supplementary material.}
\vspace{-15pt}
\label{fig4}
\end{figure}

\subsection{Loss Function Design}

\subsubsection{Normalized Cross-Correlation (NCC) Loss}

We utilize NCC \cite{yoo2009fast} to evaluate the similarity between two images within localized regions and capture structural information while maintaining invariance to brightness variations to maximize the normalized cross-correlation between corresponding local regions of the two images and enhance the alignment accuracy.

\begin{equation}
\mathcal{L}_{\text{NCC}} = 1 - \frac{1}{N} \sum_{n=1}^{N} \frac{(I_1^n - \mu_{I_1}^n)(I_2^n - \mu_{I_2}^n)}{\sigma_{I_1}^n \sigma_{I_2}^n}
\label{eq:simplified_ncc_loss}
\end{equation}
where \( N \) denotes the total number of elements. \( I^n \) represents the intensity values of the input images at the \( n \)-th position. \( \mu_{I}^n \) denotes the local mean intensities of \( I \) within the window. \( \sigma_{I}^n \) denotes the local standard deviations of \( I \).

\begin{figure*}
\centerline{\includegraphics[width=1.0\textwidth]{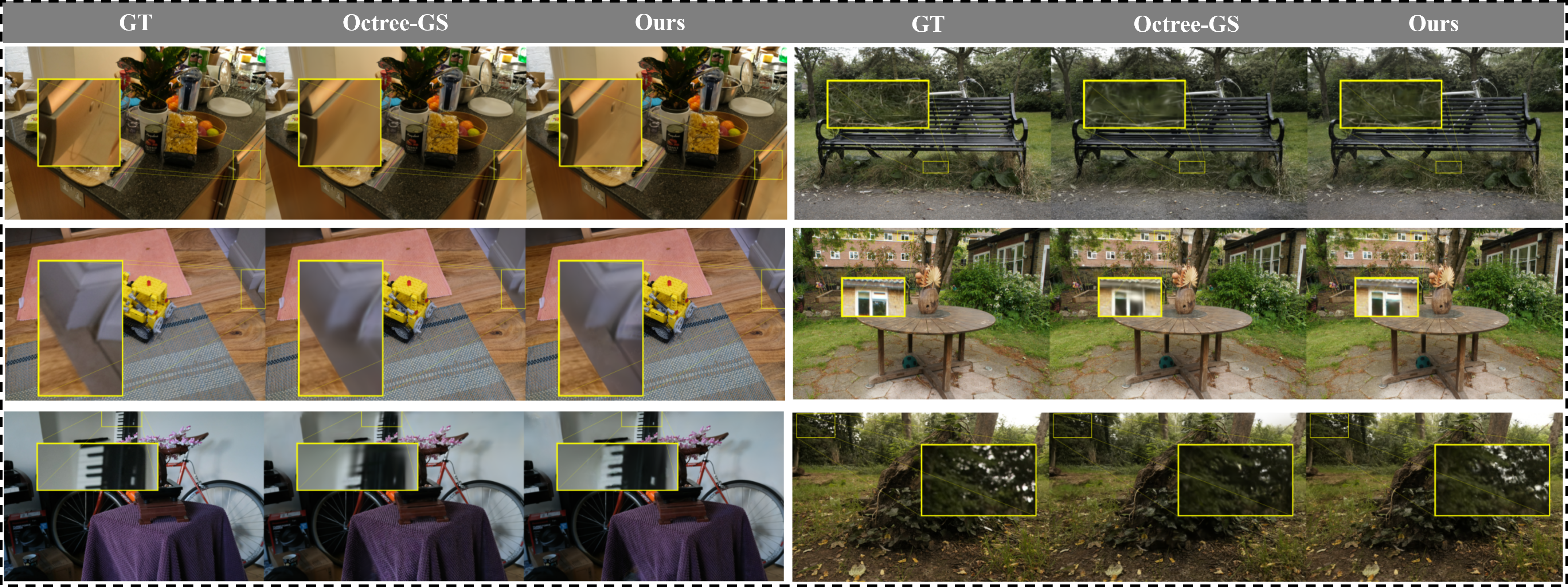}}
\caption{\textbf{Qualitative comparision of EG-Gaussian and Octree-GS.} Visual differences are highlighted with yellow insets for better clarity. Our approach consistently outperforms Octree-GS, which achieved the second-best performance on the Mip-NeRF360 dataset, demonstrating clear advantages in challenging scenarios such as thin geometries and fine-scale details. Best viewed in color. }
\label{fig3}
\vspace{-10pt}
\end{figure*}

\subsubsection{Laplacian Pyramid Loss}

To ensure that the structural details of the images are preserved across multiple scales and improve the visual quality and consistency of the aligned images, we design a Laplacian Pyramid loss which measures the structural similarity between two images across multiple scales by constructing their respective Laplacian pyramids. 

\begin{equation}
\mathcal{L}_{\text{Laplacian}} = \sum_{l=1}^{L} \| \mathcal{L}_1^{(l)} - \mathcal{L}_2^{(l)} \|_1
\label{eq:laplacian_loss}
\end{equation}
where \( L \) denotes the number of levels in the Laplacian pyramid. \( \mathcal{L}_1^{(l)} \) and \( \mathcal{L}_2^{(l)} \) are the Laplacian pyramid representations of images \( I_1 \) and \( I_2 \) at level \( l \) respectively.

\subsubsection{Pixel-based Loss}

For pixel-level rendering results, we employ $\mathcal{L}_\mathrm{1}$ and Structural Similarity Index Measure (SSIM) loss functions \cite{wang2004image} $\mathcal{L}_\mathrm{SSIM}$ and incorporate a volume regularization term \cite{lombardi2021mixture} $\mathcal{L}_{\mathrm{vol}}$ to supervise the accuracy of the rendered outputs.

\begin{equation}
\mathcal{L_\mathrm{pixel}}=\mathcal{L}_1+\lambda_\mathrm{SSIM}\mathcal{L}_\mathrm{SSIM}+\lambda_\mathrm{vol}\mathcal{L}_\mathrm{vol},
\label{eq:pixel_loss}
\end{equation}
where the volume regularization $\mathcal{L}_{\mathrm{vol}}$ is:
\begin{equation}
\mathcal{L}_{\mathrm{vol}}=\sum_{i=1}^{N_{\mathrm{ng}}}\mathrm{Prod}(s_i).
    \label{eq:vol_loss}
\end{equation}

Here, ${N_{\mathrm{ng}}}$ denotes the number of neural Gaussians in the
scene and Prod(·) is the product of the values of the scale $s_i$ of each neural Gaussian. The
volume regularization term encourages the neural Gaussians to be small with minimal overlapping.

\subsubsection{Loss for Training}
In summary, our final training loss $\mathcal{L}$ consists of the pixel-based loss $\mathcal{L_\mathrm{pixel}}$, the laplacian pyramid loss $\mathcal{L}_{\text{Laplacian}}$, the NCC loss $\mathcal{L}_{\text{NCC}}$:

\begin{equation}
\mathcal{L}=\mathcal{L}_\mathrm{pixel}+\lambda_\mathrm{\text{Laplacian}}\mathcal{L}_{\text{Laplacian}}+\lambda_\mathrm{\text{NCC}}\mathcal{L}_{\text{NCC}},
\label{eq:total_loss}
\end{equation}

\section{Experiments}
\label{sec:experiments}

We compare our method, EG-Gaussian, with current state-of-the-art scene reconstruction methods including Mip-NeRF\cite{barron2021mipnerf}, 3DGS\cite{kerbl20233d}, GOF\cite{yu2024gaussian}, PGSR\cite{chen2024pgsr}, Scaffold-GS\cite{lu2024scaffold} and Octree-GS\cite{ren2024octree}. Part of results are summarized in Fig. \ref{fig3}, Tab. \ref{tab4}, Tab. \ref{tab1}, Tab. \ref{tab3} and Tab. \ref{tab5}.

\subsection{Experimental Settings}

\subsubsection{Datasets}
We evaluate our proposed method on three benchmark datasets widely used in the 3D reconstruction community:
Mip-NeRF360~\cite{barron2022mip}, a high-resolution dataset capturing diverse indoor and outdoor scenes;
DTU~\cite{jensen2014large}, a structured dataset providing high-quality multi-view images of various objects and scenes.

\subsubsection{Implementation Details}
Both 3DGS and our method were trained for 30k iterations. For our method, we set the number of neighbor nodes \(k = 10 \) for all experiments. The two loss weights \( \lambda_\mathrm{SSIM} \) and \( \lambda_\mathrm{vol} \) are set to 0.2 and 0.001 in our experiments. For the laplacian pyramid loss, we set $\lambda_\mathrm{\text{Laplacian}}=1.0$. For the NCC loss, we set $\lambda_\mathrm{\text{NCC}}=0.01$. 

\begin{table*}
\caption{PSNR scores for DTU scenes. Our approach attains the highest signal-to-noise ratios across the DTU scenes, thereby demonstrating its capability to generate reconstructions with minimal distortion and high accuracy.}
\setlength{\tabcolsep}{3pt}
\renewcommand{\arraystretch}{1.2}
\begin{tabular}{lccccccccccccccc}
\toprule
Method   & scan24         & scan37         & scan40         & scan55         & scan63         & scan65         & scan69         & scan83         & scan97         & scan105        & scan106        & scan110        & scan114        & scan118        & scan122        \\
\midrule
3DGS \cite{kerbl20233d}         & 24.83          & \textbf{25.42}          & 25.47          & 27.52          & 28.75          & 28.05          & 26.21          &
28.49         & 26.91          & 29.71          & 33.22          & 33.13          & 30.42          & 34.49          & 33.09          \\
GOF \cite{yu2024gaussian}          & 24.97          & 25.17          & 24.36          & 27.13          & 29.37          & 29.03          & 26.63          & 28.49          & 26.44          & 27.99          & 33.01          & 31.27          & 29.42          & 33.89          & 33.39          \\
PGSR \cite{chen2024pgsr}         & 24.53          & 22.83          & 23.79          & 28.60          & 29.48          & 29.16          & 27.01          & 27.73          & 26.89          & 29.13          & 33.08          & 31.56          & 29.91          & 33.84          & 33.61          \\
Scaffold-GS\cite{lu2024scaffold}   & 27.34          & 24.59          & 26.02          & 28.32          & 28.44          & 29.67          & 27.03          & 29.54          & 27.26          & 29.84          & 33.44          & 31.86          & 30.17          & 34.12          & 34.49          \\
Octree-GS \cite{ren2024octree}    & 25.29          & 24.02          & 25.77          & 28.64          & 29.08          & 29.32          & 27.13          & 29.37          & 27.44          & \textbf{30.22}          & 33.28          & 31.55          & 29.95          & 34.31          & 33.76          \\
\midrule
\textbf{Ours} & \textbf{28.83} & 24.73 & \textbf{27.70} & \textbf{31.37} & \textbf{31.02} & \textbf{31.32} & \textbf{27.15} & \textbf{31.93} & \textbf{29.92} & 29.68 & \textbf{33.88} & \textbf{32.38} & \textbf{30.97} & \textbf{35.75} & \textbf{36.48} \\
\bottomrule
\end{tabular}
\label{tab4}
\vspace{-15pt}
\end{table*}

\subsection{Results}

Our experimental results consistently highlight the superior performance of EG-Gaussian across a comprehensive range of evaluation metrics. Notably, EG-Gaussian outperforms existing techniques in terms of geometric consistency and the preservation of fine details in rendered results based on two key techniques—epipolar geometry and graph-based spatial learning—to ensure that the generated 3D points are geometrically consistent and spatially coherent. Epipolar geometry provides strong initial constraints, while the graph-based module refines these constraints by learning spatial relationships among 3D points, which significantly improves the overall understanding of the scene's structure by capturing fine-grained spatial details.

Specifically, on the Mip-NeRF360 dataset, our model shows an exceptional ability to effectively handle complex, large-scale scenes, to preserve fine-scale details for high-quality 3D scene reconstruction. This is particularly crucial in scenarios with intricate geometric structures and sparse viewpoints, where traditional methods typically face challenges in maintaining both accuracy and consistency. 

As illustrated in the Fig. \ref{fig3}, EG-Gaussian consistently outperforms others in capturing fine-grained details, particularly in regions with thin or complex geometries. Compared to other state-of-the-art methods, our approach exhibits superior continuity and spatial coherence in the 3D reconstruction. 
The visual results reveal EG-Gaussian excels at generating high-fidelity reconstructions with minimal artifacts, even in areas with sparse data. 
In contrast, other methods often fail to preserve fine structural details or introduce inconsistencies, especially in geometrically challenging regions.

In addition, EG-Gaussian highlights its robustness in maintaining reconstruction quality in regions prone to ambiguity or detail loss in traditional methods.
For example, in geometrically complex regions, EG-Gaussian not only preserves the continuity of 3D surfaces but also accurately captures subtle variations in depth, ensuring high levels of realism and fidelity. Moreover, EG-Gaussian shows a marked improvement in handling challenging scenarios, such as occlusions and sparse data, where other models tend to generate noisy or incomplete reconstructions.

Furthermore, our approach demonstrates significant advancements in spatial consistency across varying camera perspectives. The integration of epipolar geometry and graph-based spatial learning ensures that 3D points are both geometrically constrained and spatially coherent, even in difficult reconstruction conditions. This synergy between the two components results in a more stable and reliable model, capable of producing accurate 3D reconstructions regardless of scene complexity or viewpoint coverage.

\begin{table}
\caption{average metrics scores for Mip-NeRF360 and DTU scenes. On both datasets, our method achieves the highest SSIM and PSNR scores and the lowest LPIPS score.}
\setlength{\tabcolsep}{6pt}
\renewcommand{\arraystretch}{1.2}
\scalebox{0.95}{
\begin{tabular}{l|ccc|ccc}
\toprule
\multirow{2}{*}{Method}           & \multicolumn{3}{c|}{Mip-NeRF360}                       & \multicolumn{3}{c}{DTU}                          \\  
& SSIM           & PSNR           & LPIPS          & SSIM           & PSNR           & LPIPS          \\ \midrule
3DGS\cite{kerbl20233d}           & 0.870          & 28.69          & 0.182          & 0.924          & 29.09          & 0.122          \\
GOF\cite{yu2024gaussian}            & 0.874          & 28.74          & 0.177          & 0.910          & 28.70          & 0.169          \\
PGSR\cite{chen2024pgsr}           & 0.876          & 28.56          & 0.172          & 0.910          & 28.74          & 0.178          \\
Scaffold-GS\cite{lu2024scaffold}    & 0.869          & 29.10          & 0.206          & 0.885          & 29.47          & 0.290          \\
Octree-GS\cite{ren2024octree}      & 0.867          & 29.11          & 0.188          & 0.883          & 29.27          & 0.289          \\ \midrule
\textbf{Ours}  & \textbf{0.877} & \textbf{29.72} & \textbf{0.163} & \textbf{0.941} & \textbf{30.68} & \textbf{0.090} \\ 
\bottomrule
\end{tabular}}
\label{tab1}
\vspace{-12pt}
\end{table}

\begin{table}
\caption{PSNR scores for Mip-NeRF360 scenes. Our approach achieves the highest PSNR scores in the majority of scenes, demonstrating its exceptional ability to produce low-distortion, high-fidelity 3D reconstructions.}
\setlength{\tabcolsep}{3pt}
\renewcommand{\arraystretch}{1.2}
\scalebox{0.95}{
\begin{tabular}{lccccccc}
\toprule
Method  & room                                  & bonsai                                & stump                                 & kitchen                               & counter                               & garden                                & bicycle                               \\ \midrule
3DGS \cite{kerbl20233d}        & 30.63                                 & 31.98                                 & 26.55                                 & 30.32                                 & 28.70                                 & 27.41                                 & 25.25                                 \\
GOF \cite{yu2024gaussian}         & 30.32                                 & 31.59                                 & 26.95                                 & 30.83                                 & 28.70                                 & 27.32                                 & 25.45                                 \\
PGSR\cite{chen2024pgsr}         & 30.07                                 & 31.51                                 & 26.82                                 & 30.81                                 & 28.39                                 & 27.13                                 & 25.21                                 \\
Scaffold-GS\cite{lu2024scaffold}  & 31.93                                 & 32.70                                 & 26.27                                 & 31.77                                 & 29.34                                 & 27.17                                 & 24.50                                 \\
Octree-GS \cite{ren2024octree}   & 32.35                                 & 31.84                                 & 26.32                                 & 31.14                                 & 29.52                                 & 27.54                                 & 25.04                                 \\ \midrule
\textbf{Ours}         & {\color[HTML]{000000} \textbf{32.57}} & {\color[HTML]{000000} \textbf{34.09}} & {\color[HTML]{000000} \textbf{27.11}} & {\color[HTML]{000000} \textbf{32.27}} & {\color[HTML]{000000} \textbf{30.01}} & {\color[HTML]{000000} \textbf{27.66}} & {\color[HTML]{000000} \textbf{25.57}} \\ 
\bottomrule
\end{tabular}}
\label{tab3}
\vspace{-12pt}
\end{table}

\subsection{Ablations}

We evaluate the effectiveness of both the epipolar geometry and the graph module through ablation studies. As shown in Tab. \ref{tab5}, the results confirm that both components are essential for the enhanced performance observed in our model. 

\subsubsection{Effect of Epipolar Geometry Module}

The epipolar geometry module is particularly critical when dealing with sparse viewpoint constraints, where ambiguities in 3D point localization can arise. When the epipolar geometry initialization module is removed, a significant decrease in the accuracy of 3D reconstructions is observed. 
Without epipolar constraints, the model becomes more susceptible to ambiguities in 3D point localization, especially in regions with sparse viewpoint coverage. 
In such regions, where triangulation of 3D points is particularly challenging, the absence of epipolar geometry results in errors and inconsistencies in point placement. 
This highlights the importance of epipolar geometry in constraining the potential locations of 3D points, thereby minimizing triangulation errors and reducing ambiguity in the reconstruction process. 
By enforcing geometric relationships across multiple views, epipolar geometry significantly improves the initial accuracy of 3D point estimation, particularly in complex scenes with limited data.

\subsubsection{Effect of Graph Module}

Additionally, graph module further strengthens our approach by modeling the spatial correlations between 3D points, where the nodes in the graph correspond to voxelized 3D points, with edges representing the spatial relationships between them.
Excluding the graph module leads to a substantial decline in the model’s ability to capture fine spatial correlations between 3D points. 
The graph module is specifically designed to model nonlinear spatial dependencies, which are crucial for accurately reconstructing fine-grained scene details. 
Without this component, the model struggles to refine spatial relationships between points, resulting in less detailed and less spatially coherent reconstructions. The graph module, by modeling spatial relationships between 3D points, enhances the model’s ability to preserve structural details, ensuring consistency and coherence of 3D reconstruction results, particularly in regions with intricate geometric features. 
This, in turn, leads to more accurate and visually consistent reconstructions.

\begin{table}
\caption{The ablation experiment results on Mip-NeRF360 dataset. Both components are essential for the enhanced performance observed in our model.}
\setlength{\tabcolsep}{3pt}
\renewcommand{\arraystretch}{1.2}
\scalebox{1.0}{
\begin{tabular}{lccccccc}
\toprule
Method & room  & bonsai & stump & kitchen & counter & garden & bicycle \\ \midrule
w/o epipolar & 32.21 & 32.02  & 26.44 & 31.36   & 29.34   & 27.32  & 24.75   \\
w/o graph    & 32.35 & 32.94  & 26.62 & 32.75   & 29.82   & 27.64  & 25.44   \\ \midrule
Ours         & \textbf{32.57} & \textbf{33.16}  & \textbf{27.11} & \textbf{31.94}   & \textbf{30.01}   & \textbf{27.66}  & \textbf{25.57}   \\ 
\bottomrule
\end{tabular}}
\label{tab5}
\vspace{-12pt}
\end{table}

\section{Discussion}
\label{sec:discussion}

In this work, we leverage epipolar geometry and a graph-based approach to enhance the robustness and accuracy of 3D scene reconstruction. This two-step process ensures that the generated 3D points are geometrically consistent and spatially coherent.

However, there are limitations to our approach. Although epipolar geometry is effective in constraining 3D point positions, it relies heavily on accurate camera calibration and correspondences between views. In real-world applications, misalignments or errors in camera calibration can degrade the quality of 3D reconstructions. Additionally, while the graph-based method is powerful, it may become computationally intensive as the number of 3D points increases, particularly in large-scale scenes. To address this, efficient graph construction and optimization techniques will be essential to mitigate computational challenges.

\section{Conclusion}
\label{sec:conclusion}

In this paper, we proposed a novel framework to enhance the initialization and feature learning phases of 3D scene reconstruction. 
We first leveraged epipolar geometry to improve the initialization of 3D points, providing a more reliable and geometrically constrained starting point.
Additionally, we introduced a graph-based module which dynamically learns relevant spatial dependencies among 3D points, capturing both local and global scene structures. 
Finally, we proposed a spatial encoding method that combines 3D coordinates with angular relationships between neighboring points. 
Experimental results on indoor and outdoor datasets demonstrate that our approach significantly outperforms existing methods in terms of reconstruction quality.

\section*{Acknowledgment}
This research is supported by the National Natural Science Foundation of China (No. 62406267), Tencent Rhino-Bird Focused Research Program, Guangzhou-HKUST(GZ) Joint Funding Program (Grant No.2025A03J3956 \& Grant No.2023A03J0008), the Guangzhou Municipal Science and Technology Project (No. 2025A04J4070), the Guangzhou Municipal Education Project (No. 2024312122) and Education Bureau of Guangzhou Municipality.

\bibliographystyle{ieeetr}
\bibliography{icme2025references}

\vspace{12pt}

\end{document}